%% file: main.tex
\documentclass[runningheads]{llncs}
\usepackage{makecell}
\usepackage{longtable}
\usepackage[nocompress]{cite}
\usepackage{makeidx}  % allows for indexgeneration
\usepackage{xcolor}
\usepackage{graphicx}
\usepackage{subcaption}
\usepackage[subtle]{savetrees}

\newcommand{\mypar}[1]{\smallskip\noindent\textbf{#1.}}

\usepackage{amsthm}

\newcolumntype{P}[1]{>{\centering\arraybackslash}p{#1}}
\usepackage{booktabs}
\usepackage{multirow}

\usepackage{amsmath}
\usepackage{multicol}
\usepackage{marginnote}

\begin{document}
\mainmatter              % start of the contribution
\title{xSemAD: Explainable Semantic Anomaly Detection in Event Logs Using Sequence-to-Sequence Models}
\titlerunning{Sequence-to-Sequence Model-based Explainable Semantic Anomaly Detection}  % abbreviated title (for running head)
\author{Kiran Busch\inst{1} \and Timotheus Kampik\inst{2} \and Henrik Leopold\inst{1}}
%\author{Authors removed for blind review}
%
\authorrunning{Busch et al.}   % abbreviated author list (for running head)
%Craig Chambers, Kim B. Bruce, Elisa Bertino}
%
\institute{Kühne Logistics University, Hamburg, Germany\\ \email{kiran.busch|henrik.leopold@klu.org}\\ \and SAP Signavio, Berlin, Germany \\ \email{timotheus.kampik@sap.com}}
\maketitle              % typeset the title of the contribution
% \index{Ekeland, Ivar} % entries for the author index
% \index{Temam, Roger}  % of the whole volume
% \index{Dean, Jeffrey}
\vspace{-1em}
\begin{abstract}        % give a summary of your paper

\input{sections/00_abstract}

%                         please supply keywords within your abstract
\end{abstract}

\vspace{-2.6em}
\section{Introduction}

\vspace{-.7em}
\input{sections/intro}

\section{Motivation}
\label{sec:motivation}
\input{sections/motivation}

\vspace{-.7em}
\section{Approach}
\label{sec:approach}
\input{sections/approach}

\vspace{-.7em}
\section{Experimental Evaluation}
\label{sec:evaluation}
\input{sections/evaluation}

\vspace{-.7em}
\section{Related Work}
\input{sections/relatedwork}
\label{sec:rw}

\vspace{-.7em}
\section{Conclusion and Future Work}
\label{sec:conclusion}
\input{sections/conclusion}

\vspace{1em} %\\
\mypar{Acknowledgment} Part of this research was funded by the Deutsche Forschungsgemeinschaft (DFG, German Research Foundation) – Project No. 528177077.

\bibliographystyle{splncs04}
\bibliography{main.bib}

\end{document}

%% file: sections/00_abstract.tex
The identification of undesirable behavior in event logs is an important aspect of process mining that is often addressed by anomaly detection methods. 
Traditional anomaly detection methods tend to focus on statistically rare behavior and neglect the subtle difference between rarity and undesirability. The introduction of semantic anomaly detection has opened a promising avenue by identifying semantically deviant behavior.
This work addresses a gap in semantic anomaly detection, which typically indicates the occurrence of an anomaly without explaining the nature of the anomaly. We propose xSemAD, an approach that uses a sequence-to-sequence model to go beyond pure identification and provides extended explanations. In essence, our approach learns constraints from a given process model repository and then checks whether these constraints hold in the considered event log. This approach not only helps understand the specifics of the undesired behavior, but also facilitates targeted corrective actions.
Our experiments demonstrate that our approach outperforms existing state-of-the-art semantic anomaly detection methods.

%% file: sections/intro.tex
In many domains, the execution of business processes must adhere to certain rules. For example, a bank clerk can only grant a loan to a client after performing a number of specific checks and a doctor can only prescribe certain medicine after investigating possible side effects with other drugs the patient is currently prescribed to. A common strategy to detect violations against such rules is the use of conformance checking techniques~\cite{caron2013comprehensive,rozinat2008conformance}. In essence, these techniques compare the actual behavior, as recorded by information systems, to the desired behavior that is captured using a normative process model. While such techniques have been found to be valuable for both detecting and preventing violations~\cite{munoz2016conformance}, they are only applicable if a process model specifying the normative behavior is available.

Recognizing that such normative process models are not always available in practice, several authors have introduced so-called \textit{anomaly detection techniques}. The core idea of anomaly detection techniques is to discover violations, i.e. anomalous behavior, by solely analyzing the contents of an event log. Traditional anomaly detection methods, e.g.~\cite{bezerra2013algorithms,nolle2018binet,nolle2016unsupervised}, build on the assumption that anomalous behavior is \textit{rare}, and hence take a statistical angle. More recent work~\cite{van2021natural,caspary2023does} both acknowledges and illustrates that there is difference between rarity and undesirability, and that several violations can only be identified by considering the semantics of the executed events. However, depending on the chosen strategy, existing semantic anomaly detection techniques are either very inaccurate in terms of recall or suffer from a lack of explainability, i.e., it is not transparent to the user why a trace or event pair is considered anomalous.  

In this paper, we set out to address the limitation of explainability while still providing a reasonable accuracy. To this end, we propose a novel semantic anomaly detection technique that builds on a \textit{sequence-to-sequence} (seq2seq) model. The core idea behind our technique is to fine-tune a language model based on a given process model repository. In this way, we are able to learn execution patterns that later help us to detect anomalous behavior in event logs. Our technique does not only help to understand the specifics of the undesired behavior, but
also facilitates targeted corrective actions. To demonstrate the feasibility and performance of our approach, we conduct an evaluation on a large dataset from industry and compare our approach against the state of the art.

The remainder of this paper is organized as follows. Section \ref{sec:motivation} explains the problem we are addressing in this paper and highlights the research gap. 
Section \ref{sec:approach} introduces our novel semantic anomaly detection technique. 
Section \ref{sec:evaluation} presents the results from our evaluation experiments. 
Section \ref{sec:rw} reflects on related work before Section \ref{sec:conclusion} concludes the paper.

%% file: sections/motivation.tex
To illustrate the capability and potential of semantic anomaly detection techniques, consider a simplified loan application process. A regular execution of this process starts by receiving a loan application (A). Next, the credit history of the customer is checked (B) before the application is either approved (C) or rejected (D). Depending on the decision, the customer is informed about the approval (E) or the rejection (F). In case the application is approved, the funds are disbursed (G). In either case, the case is archived (H). 
\vspace*{-1em}
\begin{figure}[bth]
    \hspace*{1cm}
    \centering
    \resizebox{0.65\textwidth}{!}{%
    \begin{tabular}{l}
     \hline \noalign{\smallskip}
     \multicolumn{1}{c}{Trace $\sigma_1$}\\
     \hline \noalign{\smallskip}
      A \quad Receive loan application \\\noalign{\smallskip} 
        C \quad Approve application \\\noalign{\smallskip}     
      B \quad Check credit history \\\noalign{\smallskip}
      E \quad Send approval \\\noalign{\smallskip}
      G \quad Disburse funds \\\noalign{\smallskip}
      \hline \noalign{\smallskip}
    \end{tabular}
    \hspace*{1cm}
    \begin{tabular}{l}
     \hline \noalign{\smallskip}
     \multicolumn{1}{c}{Trace $\sigma_2$}\\
     \hline \noalign{\smallskip}
      A \quad Receive loan application \\\noalign{\smallskip}  
      C \quad Approve application \\\noalign{\smallskip}
      D \quad Reject application\\\noalign{\smallskip}
      F \quad Send rejection \\\noalign{\smallskip}
      H \quad Archive case \\\noalign{\smallskip}
      \hline \noalign{\smallskip}
    \end{tabular}}
    \hspace*{1cm}
    \caption{Anomalous traces in a loan application process}
    \label{fig:example}
\end{figure}
\vspace*{-1.5em}
Now consider the two recorded execution traces depicted in Figure~\ref{fig:example}. Upon closer inspection, several execution anomalies can be detected by considering the labels of the recorded events. In trace $\sigma_1$, the application is approved (C) before the credit history is checked (B) and the archiving case activity (H) is missing. In trace $\sigma_2$, the application is both approved (C) and rejected (D) in the same trace. The automated detection of such anomalies can be conducted on three different levels: 
\begin{enumerate}
    \item An anomaly can be detected on the \textit{trace level}, i.e. a trace is marked as anomalous as a whole. In the example above, both $\sigma_1$ and $\sigma_2$ would be marked as anomalous respectively.  
    \item An anomaly can be detected on the \textit{event level}, i.e. a particular event or pair of events is marked as anomalous. In the example above, we would, for instance, mark the event H from $\sigma_1$ as anomalous (as it is missing) as well as the event pair C and B from $\sigma_1$. 
    %\update{R2.06}{(as it is missing)} as well as the event pair C and B from $\sigma_1$.  
    \item An anomaly can be detected on the \textit{constraint level}, i.e. a particular event or pair of events is marked as anomalous including the violated constraint. In the example above, event H from $\sigma_1$ is marked as \textit{missing} and event pair C and B are marked as violating the \textit{exclusive choice constraint}. 
\end{enumerate}

While prior work on semantic anomaly detection generally addresses all three levels, they have a number of important limitations. The work from Van der Aa et al. \cite{van2021natural} is able to detect anomalies on the constraint level, distinguishing three types of constraints: order, exclusion, and co-occurrence. It, however, can only detect anomalies for events that share the same business object. From the three violations discussed above, it therefore would only be able to correctly recognize the event pair C and D as exclusion violation, since only these events share the business object ``\textit{application}''. The more recent work from Caspary et al. \cite{caspary2023does} overcomes the limitation of the same business object, but can only detect anomalies on the event level, i.e. it cannot explain what is wrong with an anomalous event pair. By strictly focusing on events pairs, it can also not detect missing events, such as H in $\sigma_1$.

In this work, we aim to overcome these limitations of existing semantic anomaly detection techniques. We argue that a useful semantic anomaly detection technique needs to target the constraint level in order to \textit{explain} to the user what is wrong about a considered event or event pair. 
In the following we focus on control flow based constraints.
%\update{R2.02}{In the following we focus on control flow based constraints.} 
It should further be able to generalize from the behavior it was trained on and appropriately consider the context of a trace. The latter is important since an event pair that is considered anomalous in one trace might not necessarily be considered anomalous in every trace. To address these challenges, we adopt a seq2seq model and combine it with the recognition of declarative execution constraints. The next section introduces our technique in detail.

%% file: sections/approach.tex
In this section, we present the xSemAD approach for explainable semantic anomaly detection. First, Section \ref{sec:approach:overview} gives an overview of our approach. Then, Sections \ref{sec:approach:training} and \ref{sec:approach:application} explain the phases and components in detail. 

\subsection{Overview}
\label{sec:approach:overview}

The xSemAD approach consists of a training phase and a detection phase, and three core components. 

\begin{figure}[tb]
%\vspace{-2em}
    \centering
    \includegraphics[width=0.9\linewidth]{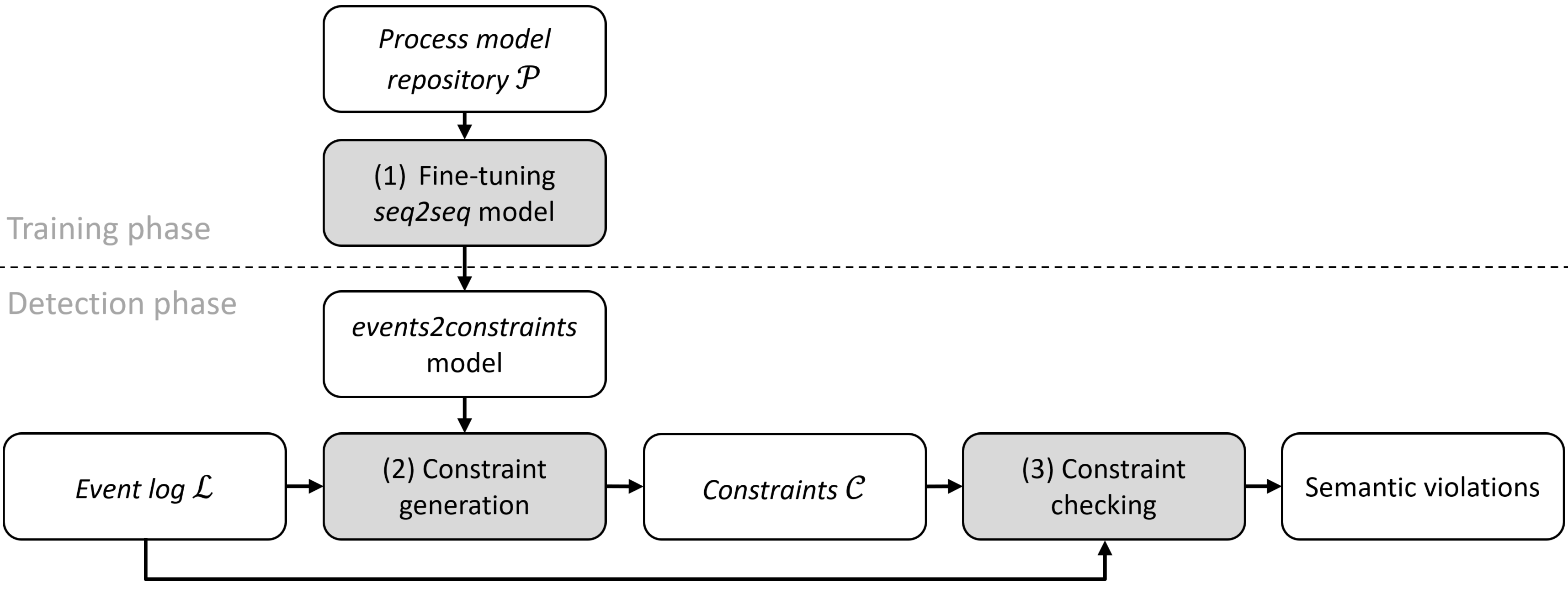}
    \caption{Overview of our xSemAD approach. }
    \label{fig:approach-overview}
\vspace{-2em}
\end{figure}

Figure \ref{fig:approach-overview} visualizes the xSemAD approach. In the training phase, the overall idea is to use an existing process model repository to learn the execution relations between activities and capture those by fine-tuning a seq2seq model (see component 1). In the anomaly detection phase, we then use the fine-tuned model, which we refer to as \textit{events2constraints} model, to determine the constraints that have to hold for a considered event log (see component 2).  To illustrate this, consider the traces $\sigma_1$ and $\sigma_2$ from Figure \ref{fig:example}. Even without knowing the order of the involved activities, we are intuitively able to determine an order for several events. Among others, we could argue that the process should always start with the ``\textit{Receive loan application}'' event or that the event ``\textit{Check credit history}'' must precede either the event ``\textit{Approve application}" or ``\textit{Reject application}''. The xSemAD approach uses the knowledge encoded in the events2constraints model to derive such constraints for the given event log. Finally, we check whether the constraints generated for the event log violate the actual execution relations in the event log (see component 3). The final outcome is, respectively, a set of semantic violations. Since the xSemAD approach is able to specify which constraints have been violated, it can use these constraints to explain the violations, thus enhancing explanatory capabilities in semantic anomaly detection.

\subsection{Training phase}
\label{sec:approach:training}

In the training phase, we use a process model repository as input to fine-tune a seq2seq model. The output is the fine-tuned model, i.e., the events2constraints model. 

\mypar{Input}
Process models define execution dependencies between the activities of a process. For the purpose of this paper, it is sufficient to abstract from specific process modeling languages, such as BPMN or Petri nets, by focusing on the relationships of activities. 
%Instead, the focus is on the behaviors described by a model, with an emphasis on examining the relationships between a set of activities. 
%We denote the universe of such activities by $\mathcal{A}$ and 
We use $\mathcal{P}$ to refer to a set of process models, i.e. a process model repository. Then, a process model $P\in \mathcal{P}$ defines a set of activity execution sequences $\Pi$ that %capture sequences of activity executions that 
lead the process to its final state. For a given execution sequence $\pi = \langle a_{1}, \dots, a_{n} \rangle \in \Pi$ of length $n$, we use $a_{j}$, $1 \leq j \leq n$, to refer to the $j^{th}$ activity in $\pi$. 

\mypar{Fine-tuning}
To capture the execution constraints that must hold between activities, we fine-tune the seq2seq model Flan-T5~\cite{flant5}, which is a pre-trained language model trained on various supervised and unsupervised tasks.
To fine-tune Flan-T5 for constraint generation, we train it using the process model repository $\mathcal{P}$, specifically focusing on input-target pairs. This results in the development of a task-specific seq2seq model, the events2constraints model. After training, the events2constraints model can be employed for checking the execution constraints of any input event log.

To create the training dataset, we first need to define the input-target pairs for each training sample. The input value of a training sample contains all activities $P_A$ of a process model $P \in  \mathcal{P}$.
The target value of a training sample is determined based on a specific constraint associated with the set $P_A$ during its training phase. To obtain these target values, we employ the declarative process language \textsc{declare}~\cite{DiCiccio2022} which provides abstractions on top of finite-trace linear temporal logic that are useful for the process domain. Below, we give an informal overview of the \textsc{declare} constraints that we use\footnote{For a more comprehensive introduction to \textsc{declare}, we refer to~\cite{DiCiccio2022}.}.
\begin{enumerate}
    \item \textit{Initiation constraint:} $\text{{Init}}(a)$ -- The process starts with activity $a$.
    \item \textit{Termination constraint:} $\text{{End}}(a)$ -- The process ends with activity $a$.
    \item \textit{Succession constraint:} $\text{{Succ}}({a_j}, {a_k})$ -- Activity ${a_j}$ (${a_k}$) occurs if and only if it is followed (preceded) by activity ${a_k}$ (${a_j}$) in the process.
    \item \textit{Alternate succession constraint:} $\text{{AltSucc}}({a_j}, {a_k})$  -- Activities ${a_j}$ and ${a_k}$ occur in the process if and only if the latter follows the former, and they alternate each other in the trace.
    \item \textit{Choice constraint:} $\text{{Ch}}({a_j}, {a_k})$ -- Activity ${a_j}$ or activity ${a_k}$ must be in the process.
    \item \textit{Co-existence constraint:} $\text{{CoEx}}({a_j}, {a_k})$ -- Both activities ${a_j}$ and ${a_k}$ must be in the process.
    \item \textit{Exclusive choice constraint:} $\text{{ExCh}}({a_j}, {a_k})$ -- Either activity ${a_j}$ or activity ${a_k}$ (but not both) must be in the process.
    \item \textit{Response constraint:} $\text{{Resp}}({a_j}, {a_k})$ -- If activity ${a_j}$ occurs in the process, then activity ${a_k}$ occurs after ${a_j}$.
    \item \textit{Alternate response constraint:} $\text{{AltResp}}({a_j}, {a_k})$ -- Each time activity ${a_j}$ occurs in the process instance, activity ${a_k}$ occurs afterwards, before ${a_j}$ recurs.
    \item \textit{Precedence constraint:} $\text{{Prec}}({a_j}, {a_k})$ -- Activity ${a_k}$ occurs in the process instance only if preceded by activity ${a_j}$.
    \item \textit{Alternate precedence constraint:} $\text{{AltPrec}}({a_j}, {a_k})$ -- Each time activity ${a_k}$ occurs in the process instance, it is preceded by activity ${a_j}$, and no other ${a_k}$ can recur in between.
\end{enumerate}

Note that while we employ \textsc{declare} in this paper, our approach can analogously be used with other constraint types, i.e., the \textsc{declare} constraint types can be exchanged with other constraint types.

For each process $P_i$ in the model repository $\mathcal{P}$, the set of constraints is given by $\mathcal{C}_i = \{C_{i_1}, C_{i_2}, \ldots, C_{i_{m_i}}\}$, where $m_i$ is the total number of constraints associated with $P_i$. Each constraint $C_{i_k}$ is categorized into a specific constraint type (e.g., initiation, choice or response), denoted as $T_{i_k}$. We define a training sample as a input-target pair $(T_{i_k}:\text{{shuffle}}(P_{i_A}), C_{i_k})$, where $1 \leq k \leq m_i$, representing the set of activities in $P_i$ and one constraint $C_{i_k}$ with type $T_{i_k}$. Here, the colon ``:'' separates the constraint type from the shuffled list of activities.
%\update{R3.05}{ Here, the colon ``:'' separates the constraint type from the shuffled list of activities.}
%selected from the set of constraints associated with $P_i$. 
We randomize the order of $(P_{i_A})$ in each training sample using $\text{{shuffle}}(P_{i_A})$ to ensure that the model remains unbiased by the order of the activity labels, forcing it to focus on learning the intrinsic relationships among these labels.
The training dataset is then defined as the union of all training samples:
$\mathcal{M}_{train} = \bigcup_{i=1}^{n} \bigcup_{k=1}^{m_i} (T_{i_k}:\text{{shuffle}}(P_{i_A}), C_{i_k}) $, where $n$ is the number of process models in $\mathcal{P}$.
%$\mathcal{M}_{train} = \bigcup_{i=1}^{n} \{(T_{i_k}:\text{{shuffle}}(P_{i_A}), C_{i_k}) \mid P_A \text{ is the set of activities in } P_i, C_{i_k} \in \mathcal{C}_i, T_{i_k} \text{ is the constraint type of } C_{i_k}\}$, where $n$ in the number of process models in $\mathcal{P}$.
We use this training dataset $\mathcal{M}_{train}$ to fine-tune Flan-T5 and thus obtain our events2constraints model. 

\mypar{Illustration} To illustrate that, consider one possible training sample with input-target pair for a simplified loan application process:
\begin{enumerate}
    %\item (\textit{Init: approve application, reject application, check credit history, receive loan application, send approval, send rejection, archive case, disburse fund; Init(receive loan application))}
    \item (\textit{Exclusive choice: archive case, reject application, check credit history, approve application, receive loan application, send approval, send rejection, disburse fund; Exclusive choice(approve application, reject application))}
\end{enumerate}
%\update{R2.03}{This training sample illustrates an ``exclusive choice'' constraint. In this context, the constraint specifies that either the activity ``approve application'' or ``reject application'' (but not both) must be present in the process execution.}

\subsection{Detection phase}
\label{sec:approach:application}

%Given a universe of events $\mathcal{E}$ that can occur during process execution, a trace $\sigma$ is a sequence of events, i.e., $\sigma = \langle e_1, \dots, e_n \rangle$. We use $e_j, 1 \leq j \leq n$, to refer to the $j^{th}$ event in $\sigma$. We further use $\lambda(e_{j})$ to refer to the label of that event. 
%Finally, an event log $L = \{\sigma_1, \dots, \sigma_n \}$ is a multiset of traces. 

The detection phase consists of the constraint generation and the constraint checking components. Below, we explain each component in detail along with a respective illustration.  

\mypar{Constraint generation}
The constraint generation component takes as input the events2constraints model and an event log $L$. The former stems from the training phase. The latter is provided by the user and will be checked for semantic anomalies. For our purposes, we consider an event log $L$ as a multiset of traces. Given the universe of events $\mathcal{E}$, a trace  $\sigma$ is defined as a sequence of events, i.e., $\sigma = \langle e_1, \dots, e_n \rangle$. We use $e_j, 1 \leq j \leq n$, to refer to the $j^{th}$ event in $\sigma$.

The first step of the constraint generation component involves the extraction of event labels from $L$. 
%Initially, the component (2) extracts a set of event labels $\mathcal{E}$, contained within $\mathcal{L}$, i.e., $\mathcal{E} = \{ e \mid e \in \bigcup_{i=1}^{m} \sigma_i, \text{ for } \sigma_i \in \mathcal{L} \}$.
In a second step, we then use these labels as input for the events2constraints model to generate constraints. 
%Notably, our model possesses the capability of autonomously generating constraints based on these event labels. This feature enhances the model's applicability to a wide range of event logs.
The input of the events2constraints model is defined by the format that we used in the training phase, i.e.,
the input sequence is composed of a constraint type and the event labels.
%The constraint generation process requires entering an input sequence into our fine-tuned model that is similar to the input value in the training sample. 
%This means that the input sequence consists of a constraint type, outlining the desired constraint type, and the corresponding event labels.

Our model employs a strategy which generates $n$ output sequences corresponding to $n$ potential constraints for a given input. This is accomplished through the use of \textit{beam search}, a decoding algorithm applied to our fine-tuned model. 
The beam search algorithm generates output sequences, i.e. constraints, auto-regressively by selecting words such that the resulting sequences have high probabilities. However, the words originate from the large vocabulary of the language model and therefore may not match the vocabulary of the event log. 
% However, one limitation of the beam search algorithm is its tendency to produce output sequences with repeating words or short patterns. 
%As a result, the output may contain event labels that are very similar to the intended names. 
To address this limitation, we refine our selection process to include only constraints with event labels that were originally found in the input sequence.
Additionally, we introduce a threshold parameter $\theta$ to further improve the output. The parameter $\theta$ is crucial in determining the set of constraints, as only those with probabilities exceeding the defined $\theta$ are considered for the output. This process enhances the precision of the generated constraints by filtering out those with lower probabilities. As a result of this procedure, the constraint generation component produces a set of possible and probable constraints for the given event log.

\mypar{Illustration} Consider an event log consisting of the traces $\sigma_1$ and $\sigma_2$ from Figure~\ref{fig:example} with the event labels ``\textit{approve application}'', ``\textit{reject application}'', ``\textit{check credit history}'', ``\textit{receive loan application}'', ``\textit{send approval}'', ``\textit{send rejection}'', ``\textit{archive case}'', and ``\textit{disburse fund}''. If the goal is to determine possible start events (i.e., the \textsc{declare} constraint \textit{Init}) based on the event labels, the corresponding input sequence for our model is \textit{(Init: approve application, reject application, check credit history, receive loan application, send approval, send rejection, archive case, disburse fund)}. 
In this example, the input sequence specifies the constraint type as \textit{Init} and provides a list of the log's unique event labels. Our model uses the beam search algorithm to generate potential constraints based on this input sequence: 
\vspace*{-1mm}
\begin{enumerate}
    \item \textit{Init(receive loan application)}
    \item \textit{Init(check credit history)}
    \item \textit{Init(receive application)}
\end{enumerate}\vspace*{-1mm}
%\update{R2.05}{Our method automatically filters out} 
Our method automatically filters out the third output sequence as it contains the event label ``\textit{receive application}'', which was not provided in the input sequence. The beam search algorithm uses conditional probabilities to track the most likely output sequences at each generation step. Suppose the probabilities for the remaining two constraints are:
%We filter out the third output sequence as it contains the event label ``\textit{receive application}'', which was not provided in the input sequence. The beam search algorithm uses conditional probabilities to track the most likely output sequences at each generation step. Suppose the probabilities for the remaining two constraints are:
\vspace*{-1mm}
\begin{enumerate}
    \item \textit{Init(receive loan application)} - Probability: 0.95
    \item \textit{Init(check credit history)} - Probability: 0.40
\end{enumerate}\vspace*{-1mm}
Given a threshold $\theta= 0.7$, the second constraint is filtered out, and the final generated constraint is \textit{Init(receive application)}, defining that the process should begin with the event ``\textit{receive application}''.
%This example demonstrates the process of generating constraints in the (2) component. 
By altering the constraint type in the input sequence, a variety of constraints $\mathcal{C}$ can be produced.

\mypar{Constraint checking}
%Following the generation of constraints in component (2), which captures the relationships between event labels, our approach progresses to the constraint checking component (3). 
In the constraint checking component, we apply the generated constraints $\mathcal{C}$ to the event log $\mathcal{L}$ to identify semantic anomalies. These anomalies then represent cases where the observed behavior deviates from the found constraints.
%The set of constraints $\mathcal{C}$ generated in the previous component is applied to the event log $\mathcal{L}$. 
To accomplish this, we check for each trace whether its sequence of events contradicts the constraints $\mathcal{C}$.
%This is done by using an anomaly detector, which checks each trace in the event log against the set of constraints to identify instances of constraint violations.
%For each trace in the event log, the anomaly detector identifies if any of the constraints have been violated. A violation occurs when the observed sequence of events in a trace contradicts the constraints established for the process. 
As a result, we obtain a set of anomalies and the specific constraints these anomalies violate. The violated constraint is important feedback to the user as it explains what exactly is considered to 
be anomalous and facilitates addressing and correcting the detected issues. Our approach is specifically designed to detect anomalies at the constraint level; however, it can be readily adapted for anomaly detection at the trace and event levels as well.

%The result is a set of instances where semantic violations have been detected.
%Then the anomaly detector provides information on which specific constraints have been violated as soon as violations are detected. 
%The final output of this phase is a set of detected semantic violations. Each instance in this set corresponds to a trace in the event log where a violation of the derived semantic constraints has been identified. The anomaly detector not only flags these violations but also associates them with the specific constraints that have been breached.

\mypar{Illustration} Again consider the traces $\sigma_1$ and $\sigma_2$ from Figure \ref{fig:example}. Further, consider the trace 
$\sigma_3 = \langle$ \textit{Check credit history (B)}, \textit{Accept application (C)}, \textit{Send approval (E)}, \textit{Disburse fund (G)} $\rangle$. 
Suppose we have already generated the \textit{Init} constraint as explained above.
%\begin{figure}[ht]
%    \hspace*{1cm}
%    \begin{center}
%    \begin{tabular}{l}
%     \hline \noalign{\smallskip}
%     \multicolumn{1}{c}{Trace $\sigma_3$}\\
%     \hline \noalign{\smallskip}
%      B \quad check credit history %\\\noalign{\smallskip} 
%        C \quad accept application %\\\noalign{\smallskip} 
%      E \quad Send approval \\\noalign{\smallskip}
%      G \quad disburse fund \\\noalign{\smallskip}
%      \hline \noalign{\smallskip}
%    \end{tabular}
%    \end{center}
%    \hfill
%    \hspace*{1cm}
%    \caption{Anomalous traces in a loan %application process}
%    \label{fig:example_trace}
%\end{figure}
In this case, the anomaly detector, using the generated constraints~$\mathcal{C}$, would identify that the trace $\sigma_{3}$ violates the \textit{Init} constraint. Specifically, it violates the constraint that the process should start with the ``\textit{receive loan application}'' event. %Therefore, the anomaly detector would output a detected semantic violation for trace $\sigma_{3}$ and indicate that it violates the constraint "Init: receive loan application."

%This explanatory capability provides stakeholders with insights into where and why the process deviates from the expected behavior, enabling a more informed and targeted approach to addressing anomalies in the system.

%% file: sections/evaluation.tex
In our experimental evaluation\footnote{We provide the source code of the implementation under this link: \url{https://github.com/KiriBu10/xSemAD}.}, we assess the performance of xSemAD and compare it to state-of-the-art approaches. We first describe the used dataset in Section~\ref{sec:dataset} and then the experimental setup in Section~\ref{sec:experimental-setup}. Finally, we present the results of our experiments in Section~\ref{sec:results}.

\subsection{Dataset}
\label{sec:dataset}
For training and evaluating our approach, we require a large number of event logs, coming with ground-truth constraints that describe how the process should be executed. As such constraints are difficult to obtain for a generic event log, we created a new dataset from the real-world process models found in the SAP Signavio Academic Models (SAP-SAM) collection~\cite{sola2022sap}, which is the largest publicly accessible collection of business process models to date. Several models included in SAP-SAM have already been published in \cite{weske_2020_3758705}. SAP-SAM comprises over one million\footnote{This number includes duplicates.} processes and other business models in various modeling notations and languages from a wide range of domains. 
We applied a series of filtering and cleaning operations to the raw dataset to improve its quality and suitability for our analysis.
Following the guidance in \cite{sola2022sap}, we excluded vendor-provided examples that are likely duplicates. Additionally, we filtered the dataset to include only models that are represented in the BPMN 2.0 notation, have English language labels, and can be converted into sound workflow nets. The latter ensures that the process models are structurally suitable for the creation of event logs. In addition to the filtering operations, we applied label cleaning techniques to improve the consistency of the models' labels. Specifically, we removed non-alphanumeric characters in the labels, handled special cases such as line breaks, converted all letters to lowercase, and removed unnecessary spaces in the labels. 
This results in a set of 38,312 process models $\mathcal{P}$. To generate the \textsc{declare} constraints from the process models, we used the converter from BPMN to constraints presented in \cite{bergman2023bpmn2constraints}.

\mypar{Dataset split} To evaluate the effectiveness of our approach in an unbiased way, we randomly split the dataset $\mathcal{P}$ into a training dataset $\mathcal{P}_{train}$,  encompassing 75\% of the models, a validation dataset $\mathcal{P}_{validation}$, consisting of 15\% of the models, and a test dataset $\mathcal{P}_{test}$, including the remaining 10\% of the models. 
We trained our approach using $\mathcal{P}_{train}$ (see Section \ref{sec:approach:training}), while $\mathcal{P}_{validation}$ played a crucial role in validating and determining the optimal configuration for the hyperparameter $\theta$ in our model. The separately held $\mathcal{P}_{test}$ was used exclusively to evaluate the performance of our approach without any influence from the training or validation process. For all process models in~$\mathcal{P}$, we parsed the set of ground-truth constraints $\mathcal{C}_{true}$. 
In addition to that, to get our test event logs $\mathcal{L}_{test}$, we converted the process models in $\mathcal{P}_{test}$ into workflow nets to automatically generate event logs using the PM4Py\cite{berti2019process} playout functionality. To create an experimental setting that is as realistic as possible, we then created a noisy version of each event log by expanding it to 1,000 traces and introducing random noise by adding, removing or swapping events in selected traces, inspired by prior works such as \cite{bezerra2013algorithms, van2021natural, bezerra2008anomaly}.

\subsection{Experimental setup}
\label{sec:experimental-setup}
\mypar{Implementation} 
We implemented our approach in Python, using the Huggingface library \cite{huggingface2020}.
To perform sequence tokenization, we used the efficient \textit{T5} tokenizer, provided by the Huggingface tokenzizer library, which is based on Unigram\cite{kudo18} in conjunction with SentencePiece\cite{kudo18b}. 
We fine-tuned \textit{FLAN-T5-small}\cite{flant5} using the Adam algorithm\cite{DBLP:journals/corr/KingmaB14} with a constant learning rate of $4 \times 10^{-5}$. 
To enhance training stability, we utilized weight decay fix\cite{loshchilov2018decoupled}.
Our fine-tuning process included 12 epochs with a batch size of 64 and an evaluation of model performance at 400-step intervals during the evaluation. 
To reduce the risk of overfitting, we employed an early stopping mechanism, terminating the fine-tuning process if the validation loss showed no improvement for 20 consecutive steps. 
The experiments were conducted utilizing two Nvidia RTX A6000 GPUs.

\mypar{Baseline} We evaluate our method by comparing it to two semantic anomaly detection and two declarative process discovery techniques.

In the domain of semantic anomaly detection, there are only two noteworthy works, namely \cite{van2021natural} and \cite{caspary2023does}. We benchmark our approach against the state-of-the-art techniques introduced by Caspary et al.~\cite{caspary2023does} that are based on BERT and SVMs, respectively. For our experiments, we adhered to the recommended hyperparameters. For the SVM-based approach, we employed an RBF kernel with a C-value of 2 and a vector size of 600. In the case of the BERT-based approach, we set the learning rate to $5 \times 10^{-5}$ with 500 warm-up steps. 

Component 2 of our approach involves transforming event logs into declarative constraints, which allows our method to be compared to any declarative process discovery technique. 
%transforms a given event log into declarative constraints, making it interchangeable with any declarative process discovery technique. 
To evaluate our approach comprehensively, we therefore also benchmark it against two established methods in this field: \textit{Declare Miner}~\cite{maggi2012efficient} and \textit{MINERful}~\cite{ciccio2015discovery}. For the Declare Miner, our experiments used the implementation described in~\cite{DonadelloRMS22}. We adhered to standard hyperparameters for our experiments: For the \textit{Declare Miner}, we set a minimum support of 0.2, an itemsets support of 0.9, and a maximum declare cardinality of 3. For \textit{MINERful}, our configuration included a support threshold of 0.95, a confidence level threshold of 0.25, and an interest factor threshold of 0.125. Here and henceforth, we refer to the four approaches as ``SVM'', ``BERT'', ``DecM'' and ``MINERful'', respectively, for the sake of simplicity.

\mypar{Evaluation procedure}
%Let us highlight two aspects regarding the evaluation process. 
Regarding the evaluation procedure, we highlight the following two aspects.

First, anomaly detection approaches are typically evaluated by assessing their performance in classifying a trace or event pair as either anomalous or not anomalous. However, we are interested in explaining undesired behavior using the defined types of constraints. Therefore, we evaluate our approach based on its performance in the constraint generation component. Specifically, we assess how well the approach performs in generating constraints that match the ground-truth constraints. The results of the constraint generation component reflect the overall performance of the approach, as the effectiveness of anomaly detection is directly dependent on the quality of the generated constraints.

Second, existing semantic anomaly detection approaches, including those from Caspary et al.\cite{caspary2023does}, are only able to recognize \textit{eventually-follows} (EvF) interrelations ``$\rightarrow$'' between pairs of event labels from an event log. We use $e_{i} \rightarrow e_{k}$ to denote that event $e_{i}$ occurs directly or indirectly before event $e_{k}$ in a trace. If the events $e_{i}$ and $e_{k}$ can follow each other in this specific order, this is referred to as a constraint $e_{i} \rightarrow e_{k}$. However, xSemAD, DecM, MINERful are able to recognize different types of EvF relationships, represented by the constraints, and are therefore more fine-grained. In order to still be able to compare these approaches with Caspary et al.'s approaches, we need to remove the fine granularity represented by the different constraints and generalize to EvF constraints. 
We achieve this by interpreting all constraints that can express an EvF relationship in an event log as EvF constraint, i.e., \textit{Response, Precedence, Succession, Alternate Succession, Alternate Precedence, Alternate Response}, and \textit{Co-Existence}\footnote{Note that for the comparison with state of the art in the EvF scenario, every \textit{Co-existence} constraint is transformed into two disjunctively chained EvF relationships. In contrast, \emph{Exclusive Choice} and \emph{Choice} constraints cannot be transformed into EvF relationships by any meaningful interpretation.} constraints are interpreted as EvF constraints.
%We combine the following constraint into one EvF constraint: 
For instance, if the constraint generation component creates the constraints $Response(e_{i},e_{k})$ and $Succession(e_{j},e_{l})$, we interpret them as two EvF constraints, namely $e_{i} \rightarrow e_{k}$ and $e_{j} \rightarrow e_{l}$. To be able to compare all approaches, we can thus compare their performance in generating EvF constraints.
%By comparing the discovered constraints with each other, we are then able to compare the two approaches. 
Note that our interpretation of the \textsc{declare} constraints in terms of EvF relationships is merely a (somewhat naive) approximation, for the sole purpose to achieve comparability. Still, as the experiments will show, even with this approximation xSemAD achieves results that are competitive and outperform the state of the art.

\mypar{Metrics}
We evaluate our approach by comparing the generated constraints $\mathcal{C}_{pred}$ with the ground-truth constraints $\mathcal{C}_{true}$ in $\mathcal{L}_{test}$ using \textit{precison}, \textit{recall}, and \textit{$F_{1}$ score}. 
Let $tp$ be constraints that are correctly generated and are present in the ground-truth constraint set, $fp$ the constraints that are incorrectly generated and are not present in the ground-truth constraint set, and $fn$ be constraints that are incorrectly not generated but are present in the ground-truth constraint set. Then precision, recall and \textit{$F_{1}$ score} are defined as 
$\textit{precision}:=\frac{tp}{tp+fp} $, $\textit{recall}:=\frac{tp}{tp + fn}$, and $\textit{$F_{1}$ score}:=\frac{2 \times  \textit{precision} \times \textit{recall}}{\textit{precision} + \textit{recall}}$.

\subsection{Results}
\label{sec:results}
This section presents the results of our evaluation experiments. First, we focus on the role of the threshold parameter $\theta$ in our xSemAD approach. Then, we conduct a comprehensive performance evaluation comparing xSemAD with state-of-the-art methods in different scenarios.

\begin{figure}[tb]
     \centering
     \begin{subfigure}[b]{0.24\textwidth}
         \centering
         \includegraphics[width=\textwidth]{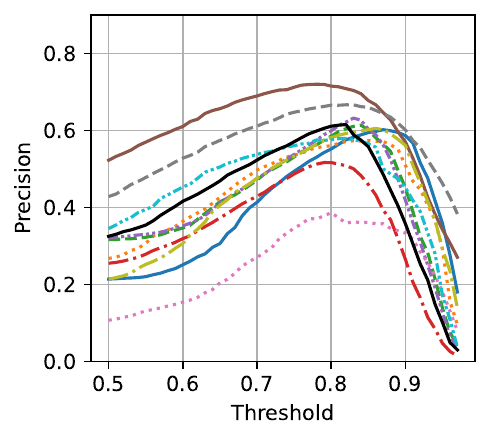}
         \caption{Precision}
     \end{subfigure}
     \hfill
     \begin{subfigure}[b]{0.24\textwidth}
         \centering
         \includegraphics[width=\textwidth]{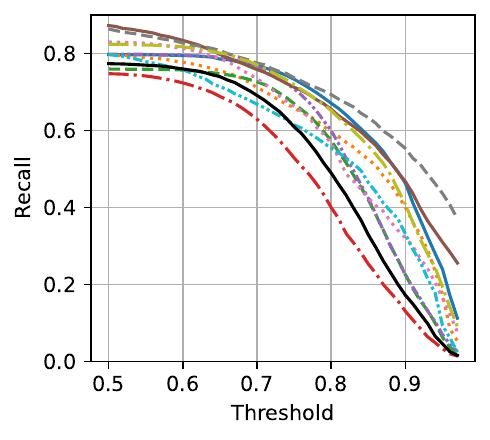}
         \caption{Recall}
     \end{subfigure}
    \hfill
     \begin{subfigure}[b]{0.24\textwidth}
         \centering
         \includegraphics[width=\textwidth]{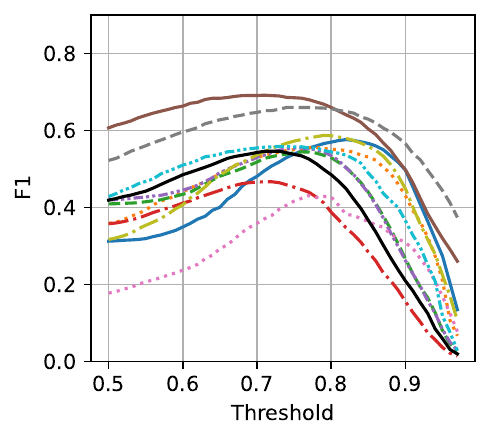}
         \caption{$F_{1}$-Measure}
     \end{subfigure}
    \begin{subfigure}[b]{0.24\textwidth}
         \centering
         \includegraphics[width=24mm]{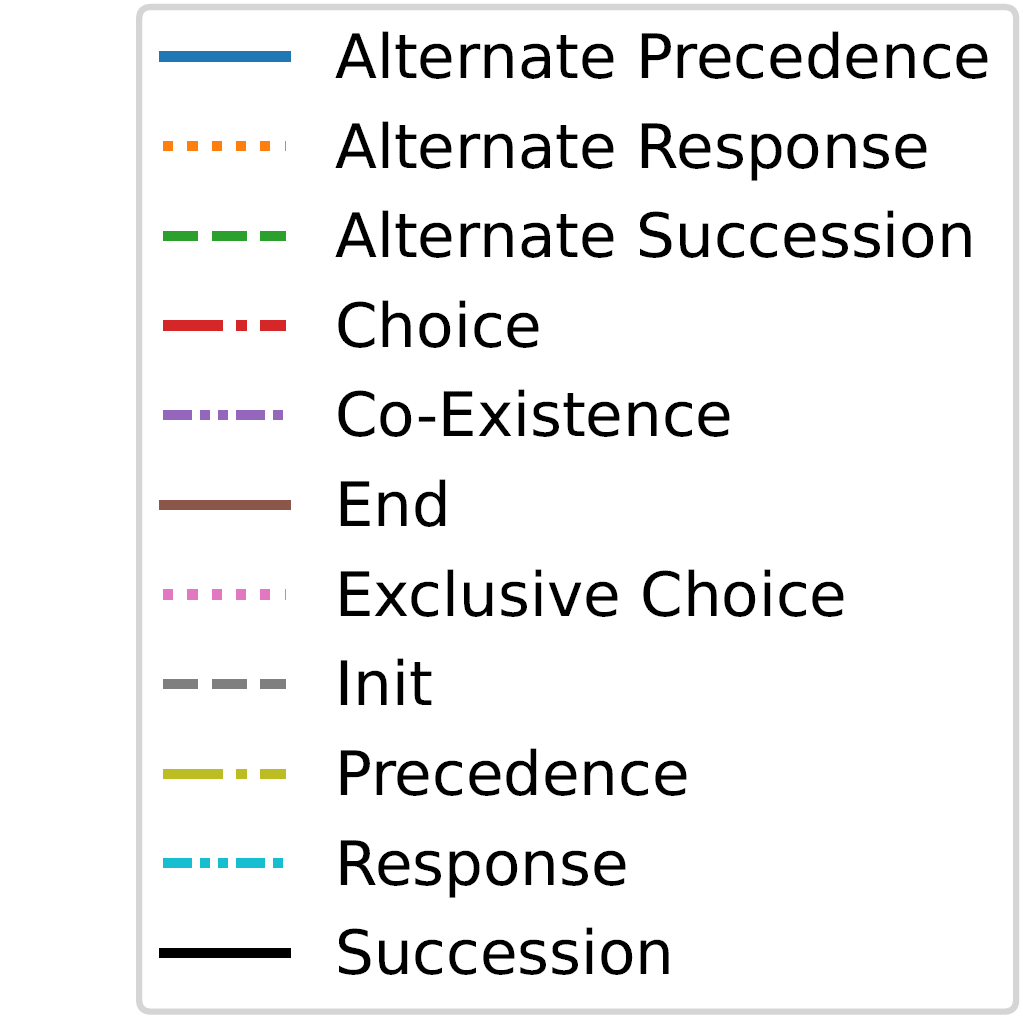}
         \vspace*{2mm}
         \caption{Legend}
     \end{subfigure}
        \caption{Precision, recall, and $F_{1}$ scores for each constraint type and different $\theta$ values}
        \label{fig:result_plots}
        \vspace{-4mm}
\end{figure}

%\mypar{Anomaly detection performance for different constraint types based on different thresholds}
\mypar{The role of the threshold parameter in xSemAD}
%To analyze the ability of our approach to detect anomalies, we compared the constraint generation performance for the different constraint types at different thresholds, 
To assess the effectiveness of our approach, we conducted a comparison of the constraint generation performance across various constraint types and thresholds $\theta$, as depicted in Figure \ref{fig:result_plots}. The x-axis represents the different $\theta$ values, while the y-axis shows the precision, recall, and $F_{1}$ scores. Each line represents a specific constraint type.
As shown in Figure \ref{fig:result_plots}, precision and $F_{1}$ scores increase for all types of constraints with an initial rise in the threshold value up to approximately $\theta=0.8$, followed by a decrease.
%As can be seen in Figure \ref{fig:result_plots}, precision and $F_{1}$ scores increase for all constraint types with an initial increase in the threshold value up to around $\theta=0.8$ and then decrease again. 
Recall decreases continuously with an increase in the threshold value, which is due to the fact that a higher threshold value leads to fewer generated constraints.
Interestingly, certain constraint types are easier to learn than others, as shown by the noticeable differences in precision, recall, and $F_{1}$ scores. In particular, the \textit{init} and \textit{end} constraints achieve the best results, suggesting that our model can effectively recognize start and end violations within process executions. However, the model has difficulties with constraint types involving split operations, such as \textit{co-existence} and \textit{choice} constraints, which have significantly lower precision and recall values. Considering the $F_1$ score, the ideal threshold value for all constraint types lies between 0.7 and 0.85.
%In summary, Figure \ref{fig:result_plots} shows that the performance in detecting violations in event logs depends on the threshold and thus the user's objective: prioritizing precision to minimize false constraint, aiming for high recall to generate more accurate constraint, or pursuing a balance between both aspects by optimizing the $F_{1}$ score.

\mypar{Comparison with state of the art}
The results presented in Table \ref{tbl:results-comparison} demonstrate the performance of xSemAD compared to state-of-the-art semantic anomaly detection and declarative process discovery methods across two different scenarios. The comparison includes the SVM and BERT approaches from Caspary et al.\cite{caspary2023does}, MINERful\cite{ciccio2015discovery}, DecM\cite{maggi2012efficient}, and our xSemAD approach. To evaluate xSemAD on the test set $\mathcal{P}_{test}$, we determined the threshold $\theta^*$ by choosing $\theta$ such that it maximizes the $F_{1}$ score on the validation set $\mathcal{P}_{validation}$. 
The table outlines the average precision, recall, and $F_{1}$ scores on $\mathcal{L}_{test}$ for each approach, providing insights on their effectiveness and differences.
The scenarios covered in the table can be broadly categorized into two groups:
\textit{(i) EvF interrelations:} In this scenario, we evaluate the ability of each approach to identify interrelations between event labels in the form of EvF constraints.
(ii)~\textit{Specific constraint:} This group of scenarios evaluates the performance of each approach in identifying different types of constraints (e.g., \emph{Init, Choice, Exclusive Choice}, etc.). 
In Section \ref{sec:motivation}, we described how anomaly detection can be performed at three different levels. Scenario (i) focuses on detecting anomalies at the event level, while scenario (ii) focuses on detecting anomalies at the constraint level. When analyzing the results across the different scenarios shown in Table \ref{tbl:results-comparison}, several important findings can be made.

First, for the EvF scenario (i), xSemAD significantly outperforms the other methods, achieving the highest precision (0.63), recall (0.78), and $F_1$ scores (0.67). This indicates xSemAD's superior capability in identifying EvF constraints, highlighting its effectiveness in capturing semantic relationships between events. The SVM and BERT approaches both record identical precision, recall, and $F_1$ scores (0.37, 0.70, and 0.46, respectively). MINERful and DecM also exhibit similar performances when compared to each other.

Second, in the detailed analysis of specific constraint types (scenario (ii)), the performance varies across different types of constraints. For initiation (\textit{Init}) and end (\textit{End}) constraints, DecM shows the best performance overall. %in precision and recall for Init, tied with xSemAD in $F_1$ score. 
However, for the \textit{End} constraint, xSemAD leads slightly in precision and matches closely in recall and $F_1$ score with DecM. This underscores the strength of xSemAD and DecM in capturing the beginning and end characteristics of event logs. 
xSemAD consistently outperforms other approaches on more complex constraints, i.e., Response (\textit{Resp}), Alternate Response (\textit{AltResp}), Succession (\textit{Succ}), Alternate Succession (\textit{AltSucc}), Precedence (\textit{Prec}), Alternate Precedence (\textit{AltPrec}), Choice (\textit{Ch}), Exclusive Choice (\textit{ExCh}) and Co-Existence (\textit{CoEx}). Overall, it achieves the highest values for precision, recall and $F_1$ for these complex constraint types, demonstrating its comprehensive ability to model various relationships between events. 

Third, comparing the performance of MINERful, DecM, and xSemAD in generating complex constraints, we observe that MINERful and DecM often achieve high recall at the expense of precision. This indicates their tendency to over-generalize, identifying many constraints as relevant, which leads to a higher number of false positives. xSemAD, however, demonstrates a balanced performance with consistently high precision and recall across various constraint types, showing its effectiveness in accurately identifying true constraints without over-generalizing.
The difference in performance can be partly attributed to the hyperparameter settings of MINERful and DecM, which influence the precision-recall trade-off. Adjusting these parameters might improve one metric at the cost of the other. As a result, it is unlikely that MINERful and DecM with other hyperparameters will achieve better results than xSemAD.  

In summary, while MINERful and DecM might show promise in recall, their low precision in complex constraint types highlights a significant limitation. In contrast, xSemAD's consistent and balanced performance underscores its superior capability in semantic anomaly detection, suggesting that even with optimal hyperparameter adjustments, MINERful and DecM are unlikely to match xSemAD's overall effectiveness in accurately detecting and explaining anomalies in process executions.

\begin{table}[!tb]
\begin{center}
\scriptsize
\begin{tabular}{ c l  c  c  c c  c  c c  c  c c  c  c c  c  l}
\toprule
%\hline
&\textbf{Scenario} & \multicolumn{3}{c}{\textbf{SVM}\cite{caspary2023does}} & \multicolumn{3}{c}{\textbf{BERT}\cite{caspary2023does}}& \multicolumn{3}{ c }{\textbf{MINERful}\cite{ciccio2015discovery}}& \multicolumn{3}{ c }{\textbf{DecM}\cite{maggi2012efficient}}& \multicolumn{3}{ c }{\textbf{xSemAD($\theta^*)$}} \\ 
%\cline{2-5}
& & Prec. & Rec. & $F_1$& Prec. & Rec. & $F_1$& Prec. & Rec. & $F_1$& Prec. & Rec. & $F_1$ & Prec. & Rec. & $F_1$ \\
\midrule

(i)&{EvF} &0.37&0.70&0.46 & 0.37&0.70&0.46 & 0.37&0.69&0.46 & 0.37&0.69&0.46& \textbf{0.63}&\textbf{0.78}&\textbf{0.67}\\ 
\midrule

(ii)&Init& - & - & - & - & - & - &0.20 & 0.85 & 0.31 & \textbf{0.76} &\textbf{0.81} &\textbf{0.76}& 0.63 & 0.76 & 0.66\\ 

&End& - & - & - & - & - & - &0.24 & 0.84 & 0.35 &0.69 &\textbf{0.76}& \textbf{0.70}& \textbf{0.70} & 0.73 & 0.69\\ 

&Resp& - & - & - & - & - & - &0.07 & 0.68 & 0.12 &0.10&\textbf{0.68}&0.16& \textbf{0.55} & 0.64 & \textbf{0.56}\\ 
&AltResp& - & - & - & - & - & - &0.07 & \textbf{0.68} & 0.12 &0.02& 0.06& 0.02& \textbf{0.52} & \textbf{0.68} & \textbf{0.54}\\ 
&Succ & - & - & - & - & - & - & 0.14 & \textbf{0.68} & 0.22 & - & - & - &\textbf{0.55} & 0.65 & \textbf{0.55}\\ 
&AltSucc & - & - & - & - & - & - &0.14 & 0.68 & 0.22 & - & - & - &\textbf{0.51} & \textbf{0.70} & \textbf{0.54}\\ 
&Prec & - & - & - & - & - & - &   0.09&0.70&0.14 &0.13 &0.70 &0.20& \textbf{0.51} & \textbf{0.74} & \textbf{0.56}\\ 
&AltPrec & - & - & - & - & - & - & 0.09&0.70&0.14 &0.03 &0.06& 0.03& \textbf{0.47} & \textbf{0.75} & \textbf{0.52}\\ 
&Ch & - & - & - & - & - & - &- & - & - &0.15 &\textbf{0.70}& 0.24& \textbf{0.47} & 0.57 & \textbf{0.46}\\ 
&ExCh & - & - & - & - & - & - &- & - & - & - & - & - &\textbf{0.43} & \textbf{0.69} & \textbf{0.48}\\ 
&CoEx  & - & - & - & - & - & - & 0.31 & 0.68 & 0.39&- & - & - & \textbf{0.58} & \textbf{0.74} & \textbf{0.60}\\ 
\bottomrule
\end{tabular}
\end{center}
\caption{Performance comparison of SVM, BERT, Minerful, DecM, and xSemAD across different scenarios on $\mathcal{L}_{test}$.}
\label{tbl:results-comparison}
\vspace{-8mm}
\end{table}

\mypar{Generalizability of xSemAD}
To address concerns of circularity—where a model's performance is artificially inflated due to its familiarity with the training data— we evaluated the robustness of xSemAD on a test set $\mathcal{P^*}_{test}\subset\mathcal{P}_{test}$ consisting of 355 models, which comprises entirely unseen process model labels, distinct from those in the training and validation sets, i.e., the labels in $\mathcal{P}_{train}\cup\mathcal{P}_{validation}$ and the labels in $\mathcal{P^*}_{test}$ have no overlap. This methodological rigor ensures that the performance reflects the actual ability of the model to generalize and not just the characteristics of the training data.

In general, this scenario represents a considerable challenge for anomaly detection approaches, as they face detection situations that are completely unfamiliar to them. If xSemAD has to generate constraints for labels that it has not yet seen in training, it is less confident in constraint generation, so that the generated constraints have lower probabilities. In this scenario, it may thus make sense to choose a less restrictive threshold. Therefore, in addition to the threshold $\theta^*$=0.73 determined on the validation set, we also tested the model with the heuristically determined threshold $\theta = 0.65$.

While the results on the unseen data are naturally lower, due to the challenging nature of this scenario, the results, shown in Table \ref{tbl:results-comparison-unseen}, confirm that xSemAD effectively generalizes from the training set. 
It demonstrates superior performance compared to SVM, BERT, MINERful, and DecM, particularly in scenarios assessing EvF interrelations and various specific constraint types. %With an $F_1$ score of 0.58 at $\theta^*=0.73$ and 0.57 at $\theta$ =0.65, 
Overall, the experiments in this scenario validate xSemAD's capabilities in detecting complex semantic anomalies, establishing its efficacy for industrial applications where model adaptability is crucial.
\begin{table}[!tb]
\begin{center}
\scriptsize
\begin{tabular}{ c l  c  c  c c  c  c c  c  c c  c  l c  c  l}
\toprule
%\hline
&\textbf{Scenario} & \multicolumn{3}{c}{\textbf{SVM/BERT}\cite{caspary2023does}} & \multicolumn{3}{ c }{\textbf{MINERful}\cite{ciccio2015discovery}}& \multicolumn{3}{ c }{\textbf{DecM}\cite{maggi2012efficient}}& \multicolumn{3}{ c }{\textbf{xSemAD}($\theta^*$)}& \multicolumn{3}{ c }{\textbf{xSemAD}($\theta=0.65$)} \\ 
%\cline{2-5}
& &  Prec. & Rec. & $F_1$& Prec. & Rec. & $F_1$& Prec. & Rec. & $F_1$& Prec. & Rec. & $F_1$ & Prec. & Rec. & $F_1$ \\
\midrule

(i)&EvF  & 0.32&0.61&0.40 & 0.32&0.60&0.40 & 0.32&0.60&0.40& \textbf{0.53}&0.74&\textbf{0.58}& 0.50&\textbf{0.79}&0.57\\ 
\midrule
(ii)&Init&  - & - & - &0.20 & \textbf{0.79} & 0.31 & \textbf{0.70} &0.77 &\textbf{0.71}& 0.37 & 0.60 & 0.43 & 0.36  &  0.71 & 0.45\\ 

&End&  - & - & - &0.24 & 0.72 & 0.34 &0.58 &0.64& \textbf{0.59}& \textbf{0.50} & 0.57 & 0.49 & 0.45 &\textbf{0.68} &0.50\\ 

&Resp&  - & - & - &0.05 & \textbf{0.50} & 0.09 &0.08&\textbf{0.50}&0.12& \textbf{0.22} & 0.36 & \textbf{0.25} & 0.19 & 0.47 & \textbf{0.25}\\ 
&AltResp&  - & - & - &0.05 & 0.50 & 0.09 &0.01& 0.03& 0.01& \textbf{0.20} & 0.39 & \textbf{0.24} &0.18 &  \textbf{0.52} & \textbf{0.24}\\ 
&Succ &  - & - & - & 0.12 & \textbf{0.60} & 0.18 & - & - & - &\textbf{0.25} & 0.49 & 0.29 & 0.23&0.58&\textbf{0.30}\\ 
&AltSucc &  - & - & - &0.12 & \textbf{0.60} & 0.18 & - & - & - &\textbf{0.24} & 0.53 & \textbf{0.29} &0.22 & 0.59 & 0.28\\ 
&Prec &  - & - & - &   0.10&0.62 &0.16 &0.14 &0.61 &0.21& \textbf{0.26} & 0.59 & \textbf{0.32} &0.21 & \textbf{0.67} &0.30\\ 
&AltPrec &  - & - & - & 0.10&0.62&0.16 &0.02 &0.04& 0.02& \textbf{0.24} & 0.61 & \textbf{0.31} &  0.21 &\textbf{0.66} & 0.29\\ 
&Ch &  - & - & - &- & - & - &0.14 &0.59& 0.22& \textbf{0.29} & 0.54 & \textbf{0.34} & 0.25 &\textbf{0.63} &0.33\\ 
&ExCh &  - & - & - &- & - & - & - & - & - &\textbf{0.25} & 0.59 & \textbf{0.31} & 0.18  & \textbf{0.70} & 0.26\\ 
&CoEx  &  - & - & - & 0.28 & 0.61 & 0.35&- & - & - & \textbf{0.43} & 0.68 & \textbf{0.47}&0.39&\textbf{0.73}&0.46\\ 
\bottomrule
\end{tabular}
\end{center}
\caption{Performance comparison of SVM, BERT, Minerful, DecM, and xSemAD across different scenarios on only unseen data.}
\label{tbl:results-comparison-unseen}
\vspace{-5mm}
\end{table}

\begin{table}[bt]
\begin{center}
\scriptsize
\begin{tabular}{ c l  c  c  c c c}
\toprule
%\hline
&\textbf{} & \multicolumn{1}{c}{\textbf{SVM}\cite{caspary2023does}} & \multicolumn{1}{c}{\textbf{BERT}\cite{caspary2023does}}& \multicolumn{1}{ c }{\textbf{MINERful}\cite{ciccio2015discovery}}& \multicolumn{1}{ c }{\textbf{DecM}\cite{maggi2012efficient}}& \multicolumn{1}{ c }{\textbf{xSemAD}} \\ 
%\cline{2-5}
\midrule
&Training time (min) & 3.4 & 274.8 & - & - & 13620.1\\ 
\midrule
&Inference time (sec)& 11.23 & 2.98 & 1.91 & 17.26 & 3.60\\ 
\bottomrule
\end{tabular}
\end{center}
\caption{Training and average inference times of our approach and the baselines.}
\label{tbl:results-comparison-computation-time}
\vspace{-10mm}
\end{table}
\mypar{Computation time}
Table \ref{tbl:results-comparison-computation-time} outlines the training and inference times for xSemAD compared to the baseline models. \emph{Training time} is defined as the duration required to train the model using $\mathcal{P}_{train}$, excluding load and preprocessing times. \emph{Inference time} is defined as the average time it takes to generate the constraints per event log in $\mathcal{L}_{test}$. The SVM-based model requires 3.4 minutes for training, while the BERT-based model takes approximately 274.8 minutes (about 4.5 hours) to fine-tune over three epochs. MINERful and DecM do not involve a training phase. xSemAD requires a significantly longer training time of 13620.1 minutes (approximately 227 hours); however, this is a one-time process, after which any number of event logs---even in several process domains and organizations---can be analyzed. It is important to asses per use case whether the substantial performance improvement justifies the extended training time. Also, since our implementation is prototypical, it can be assumed that the training time can be shortened considerably for an application in practice, e.g., by optimizing the learning rate.
Regarding inference time, all models demonstrate rapid performance, with xSemAD averaging 3.6 seconds per constraints generation of logs.

\mypar{Real-world application}
We also applied our approach to the international declarations log\cite{real-world-log} from the BPI 2020 challenge, encompassing 6449 traces and 72151 events related to travel expense claims. Although there is no gold standard available to indicate true anomalies in this process, we conducted a qualitative assessment using a similar approach to \cite{caspary2023does, van2021natural}.
Table \ref{tbl:results-real-world} focuses on key constraints that highlight notable aspects of the process. For instance, $a_1$ highlights the need for approvals before the trip concludes. $a_2$ shows a corrective mechanism where rejected declarations are eventually approved after adjustments. $a_3$ and $a_4$ confirm the necessity of approvals before payment requests. In summary, this analysis reveals critical insights into the process flow of travel expense claims, highlighting the importance of approvals, sequential dependencies, and potential areas for process improvement.
\begin{table}[!tb]
\begin{center}
\scriptsize
\begin{tabular}{ l l c }
\toprule
%\hline
\textbf{id} & \textbf{detected anomalies by xSemAD} & \textbf{Frequency} \\ 
%\cline{2-5}
\midrule
$a_1$ & Alternate Response(declaration approved by supervisor, end trip) & 9 \\ 
$a_2$ & Response(declaration rejected by supervisor, declaration approved by supervisor)& 114 \\ 
$a_3$ & Succession(declaration approved by supervisor, request payment)& 244 \\ 
$a_4$ & Alternate Precedence(request payment, declaration approved by supervisor)& 246 \\ 
\bottomrule
\end{tabular}
\end{center}
\caption{Selected anomalies detected for the  \textit{international declarations log}\cite{real-world-log}.}
\label{tbl:results-real-world}
\vspace{-8mm}
\end{table}

%% file: sections/relatedwork.tex
%Anomaly detection is an important aspect of data mining that aims to identify patterns or behaviors that deviate significantly from the norm. Two basic concepts have emerged: frequency-based anomaly detection, which relies on statistical irregularities, and semantic-anomaly detection, which focuses on identifying behaviors that stand out from a semantic perspective. 

Our research is at the intersection of anomaly detection and declarative process discovery, with a particular focus on the extraction of declarative process models from event logs in the latter domain.

\textit{Declarative process discovery} has been significantly advanced by methods such as Declare Miner\cite{maggi2012efficient}, which initially employed a brute-force strategy. It evolved through significant enhancements \cite{maggi2018parallel, maggi2011monitoring} inspired by the Apriori algorithm and the adoption of parallel processing techniques. Following Declare Miner, MINERful\cite{ciccio2015discovery} emerged as a noteworthy advancement, introducing substantial efficiency improvements. Later it has been extended with features for better efficiency, and accuracy. This includes adding flexibility through target-branched constraints\cite{di2016efficient}, streamlining by cutting redundancies\cite{di2017resolving}, and focusing on impactful constraints by removing non-meaningful ones\cite{di2018relevance}.

In the area of anomaly detection, the task of identifying undesirable behavior in the absence of normative process models is of crucial importance. This area of research can be roughly divided into frequency-based and semantic approaches.    

\textit{Frequency-based approaches} 
%are of central importance in the area of anomaly detection. These methods 
rely on statistical analyses to detect events that are rare or deviate significantly from established patterns within a data set. 
Examples are based on autoencoders \cite{krajsic2021semi, nolle2018analyzing},
LSTM \cite{krajsic2021semi} or sampling \cite{bezerra2013algorithms}.
One of the state-of-the-art methods in this area is BINet\cite{nolle2018binet}, which is a deep learning approach that uses a neural network architecture for anomaly detection. As the authors demonstrate, BINet outperforms other available methods like approaches based on sliding windows~\cite{warrender1999detecting}, SVMs, denoising autoencoders~\cite{nolle2016unsupervised}, and likelihood graphs~\cite{bohmer2016multi}. 
However, it is crucial to note that rarity detected by these methods does not necessarily mean undesirability. Recognizing that infrequency alone does not cover the spectrum of anomalies led to the development of the concept of semantic anomaly detection. In this paper, we present a novel, explainable approach targeting the constraint level. Nevertheless, we consider frequency-based approaches as complementary to ours, since frequency-based approaches tend to find different types of anomalies.  

\textit{Semantic approaches} involve identifying behavior that stands out from a semantic perspective, considering the meaning or context of the event labels.
As the first approach of its kind, Van der Aa et al. \cite{van2021natural} proposed a rule-based method for identifying undesirable behavior in event logs. However, a limitation of this method is its ability to only detect anomalies within label pairs that share the same business object.
Addressing this limitation, Caspary et al.\cite{caspary2023does} proposed two alternative approaches—based on BERT and SVMs—to enhance semantic-anomaly detection. These methods offer a broader scope by enabling the detection of semantic anomalies beyond label pairs associated with a common business object. This approach, however, can only detect anomalies on the event level.  As discussed earlier, we set out to overcome the limitations of these approaches. We argue that semantic anomaly detection technique need to provide explainable results, i.e. target the constraint level. Our approach is further able to generalize from the behavior it was trained as it considers the context of a trace.

%% file: sections/conclusion.tex
In this paper, we introduced xSemAD, which is the first semantic anomaly detection approach that can effectively target the trace, event, and constraint level without imposing limiting assumptions on the input event log. Through the use of a fine-tuned seq2seq model for constraint generation, it provides explanatory capability.
%xSemAD is the first approach that goes beyond identifying anomalies on a trace or event level by providing explanatory capability through the utilization of a fine-tuned seq2seq model. 
The core idea of xSemAD is to take an existing process model repository as input and use it to fine-tune the seq2seq model. In this way, the fine-tuned model learns constraints, i.e., in which order certain activities should be executed. By then checking to what extent these constraints are respected by a given event log, we are able to detect semantic anomalies. 
Our approach demonstrated its effectiveness through extensive evaluation on a large industry dataset, effectively generating relevant constraints and outperforming existing state-of-the-art methods such as SVM, BERT, MINERful, and Declare Miner.
While our paper addresses anomaly detection in event logs, it also represents a first step towards the discovery of temporal patterns from event logs using seq2seq models.
%However, against this background, we believe that xSemAD still represents a substantially advances the field of semantic anomaly detection. xSemAD is the first semantic anomaly detection approach that can effectively target the trace, event, and constraint level without imposing limiting assumptions on the input event log.  

However, our research is not without limitation. 
One potential limitation is the dependency on the quality and comprehensiveness of the process model repository used for fine-tuning. If the repository does not cover a wide range of possible process variations, the model may not learn all relevant constraints, leading to incomplete anomaly detection. Another limitation is the computational cost associated with training the model. It is important to consider whether the performance justifies the extended training period.
In addition, the overlap of xSemAD with the areas of temporal pattern discovery and declarative constraints leads to several critical considerations that we want to address in future work. To further increase effectiveness and accuracy, future efforts will focus on ensuring that the constraints generated by xSemAD preserve temporal semantics and do not lead to the formulation of infeasible constraints. Furthermore, exploring xSemAD's adaptability to different types of constraints beyond the specific \textsc{declare} constraints could provide deeper insights into the behavior of complex processes. 
The use of large language models also opens up several possibilities for incorporating the data perspective into the detection of anomalies.
In addition, a promising avenue for future research could be a hybrid approach, where frequency-based and semantic-based approaches complement each other. While frequency-based approaches for declarative process mining or anomaly detection completely rely on patterns in the data, semantic approaches leverage semantic knowledge that goes beyond the given data. Combing these approaches could lead to a more holistic data understanding and improved anomaly detection.